# Mechanical Orthosis Mechanism to Facilitate the Extension of the Leg

Gabrielle Lemire[1,2], Thierry Laliberté[1], Alexandre Campeau-Lecours[1,2]

[1]*Université Laval,* [2]*Centre Interdisciplinaire de recherche en réadaptation et intégration sociale (CIRRIS, Québec)*

## ABSTRACT

This paper presents the design of a mechanism to help people using mechanical orthoses to extend their legs easily and lock the knee joint of their orthosis. Many people living with spinal cord injury are living with paraplegia and thus need orthoses to be able to stand and walk. Mechanical orthoses are common types of orthopedic devices. This paper proposes to add a mechanism that creates a lever arm just below the knee joint to help the user extend the leg. The development of the mechanism is first presented, then followed by the results of the tests that were conducted.

## INTRODUCTION

In the United States, there are 282,000 persons living with a spinal cord injury, with approximately 17,000 new cases every year (54 per million persons). 41.3 % of spinal cord injuries result in an incomplete or complete paraplegia; the rest results in incomplete or complete tetraplegia [1]. The main limitation that people living with paraplegia report is being unable to walk; they also feel social pressure for standing up and walking [2]. They generally communicate that they want to be able to stand up and walk, even if it requires invasive procedures like surgery [3]. Furthermore, important physiological effects can result from remaining seated during extended periods (e.g., muscular atrophy, muscle spasticity, impaired lymphatic and vascular circulation, reduced respiratory and cardiovascular capacities [4] and sores).

Over the last years, many assistive technologies were developed to help people living with spinal cord injury stand and walk. These can mainly be separated in three categories: mechanical orthoses, functional electric stimulation (FES) orthoses and active orthoses such as exoskeletons [2]. In the last few years, there has been an effervescence in research for the development of the last two categories with the design of exoskeletons, both for assistance and for physical rehabilitation. However, even if these last two categories have a huge potential, they are often very expensive and cumbersome. The first category (mechanical orthoses) is thus mostly used in practice. This paper proposes an improvement for mechanical orthoses. The types of mechanical orthoses are Swivel Walker, the Parawalker (hip guidance orthosis), the reciprocating gait orthosis, the advanced reciprocating gait orthosis, the knee-angle-foot orthosis and the ankle-foot orthosis [2].

With some of these mechanical devices, such as the knee-angle-foot orthosis and the reciprocating gait orthosis, users need to extend the leg straight to lock the orthosis's knee joint. When the orthosis's knee joint is locked, the orthosis supports the leg so the user can stand and walk. He/she can then sit (with the legs extended) and unlock the orthosis knee joint to bend the knees. The challenge is that the orthosis needs to be extended manually by the user to lock the mechanism, which is constraining and non-ergonomic as it requires flexibility. This may cause difficulties to some users [2].

## OBJECTIVE

The objective of this paper is to modify the design of a mechanical orthosis to help users extend the leg and lock the knee in a fully extended position.

## DEVELOPMENT

In this paper, a knee-ankle-foot orthosis is used to explain the developments and validate the principle; it could then be applied to other mechanical orthoses. As explained in the introduction, in order to lock the mechanism to stand up, the users must align their thighs with their shins by pulling on the lower part of their orthosis and pushing on their thighs, as they cannot control their lower limbs. The difficulty resides in the fact that, in order for the users to pull their shins so the orthosis rotates around their knees, the force applied on the shins must be perpendicular to the latter. However, because of the orientation of the body in a sitting position, it is very difficult to do so (both for the ability to bend the body and to apply a considerable perpendicular force in that position). Figure 1a shows the force applied by the user on the shin with the lever arm $L_1$ and the two components of the force $F_{1x}$ and $F_{1y}$. Users shall thus apply a force as perpendicular as possible to the orthosis, provided only a portion of the force they apply will turn their shins (only the component of the force that is perpendicular to the shins). In Figure 1a, only the component $F_{1x}$ of the force $F_1$ helps in the rotation. Since the force applied is far from being perpendicular, users must apply a great amount of force (see Figure 1). Furthermore, the friction between their feet and the ground often increases the required amount of force. In addition, for some users, limb spasticity can sometimes require an even greater force to overcome.

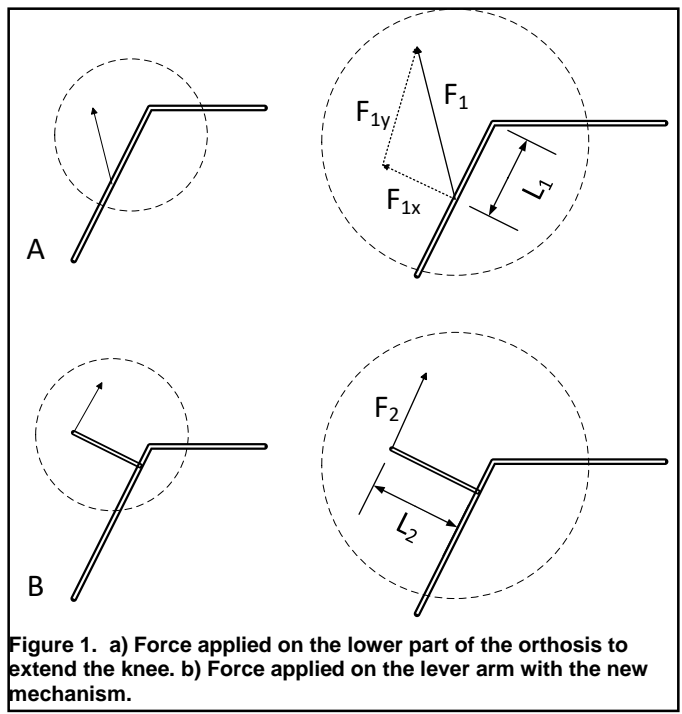

Figure 1. a) Force applied on the lower part of the orthosis to extend the knee. b) Force applied on the lever arm with the new mechanism.

This paper proposes the use of a lever arm located on the orthosis, below the knee joint (see Figure 1b). In order to align the shins and the thighs to lock the mechanism, the user would pull on the lever arm. This will produce a torque around the knee joint bringing the lower part of the orthosis to align with the upper part, and locking the mechanism for the user to be able to stand up. The longer the lever arm, the less demanding it is to align and lock the mechanism. Experiments conducted with one subject living with a spinal cord injury using a knee-ankle-foot orthosis have led to approximate the length required to lock the mechanism with little force at 20 cm. This length can be adapted to every individual.

Indeed, the proposed lever arm in Figure 1b can be fairly cumbersome, and can cause physical injuries in case of fall. A second iteration of the lever arm was thus designed to include a foldable joint. This version is in fact a knee joint without a locking mechanism. The range of motion of the joint extends from 0 to 180 degrees; the absence of lock allows the joint to move freely. However, at 0 degree, the mechanism stops and the user can apply a force upward. At rest, the link aligns with gravity (it normally aligns with the shins) and a hook and loop strip can be used to maintain it in position.

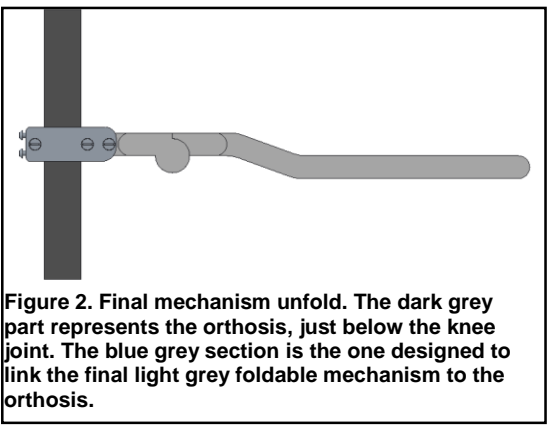

Figure 2. Final mechanism unfold. The dark grey part represents the orthosis, just below the knee joint. The blue grey section is the one designed to link the final light grey foldable mechanism to the orthosis.

To attach the mechanism on the orthosis, a versatile attachment was designed to fit on many orthoses as shown in Figure 2 and Figure 3. The three screws on the side link the mechanism to the orthosis. Also, the two screws on the back apply pressure on the back side of the orthosis so that the mechanism is always in the desired orientation. This is convenient for orthoses designed with a curved bar under the knee joint. The difference in the total length of the mechanism, fold or unfold, can be seen in Figure 2 and Figure 3. When the user does not need to use the mechanism, the bar is folded and parallel to the orthosis.

## RESULTS

The mechanism was tested with a user of knee-ankle-foot orthoses. The user found it hard to lock the orthosis without the mechanism on her own. After adding the new mechanism to her orthosis, she described it as "simple but effective, and less demanding."

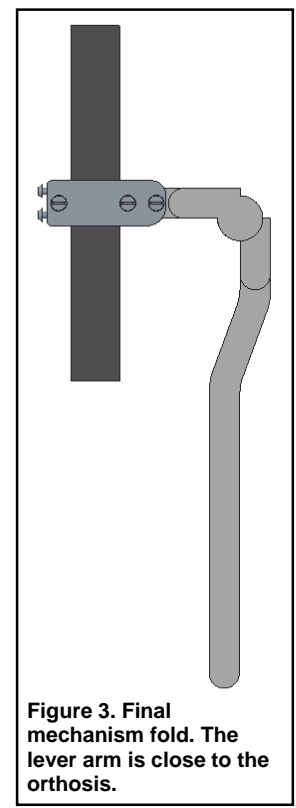

Figure 3. Final mechanism fold. The lever arm is close to the orthosis.

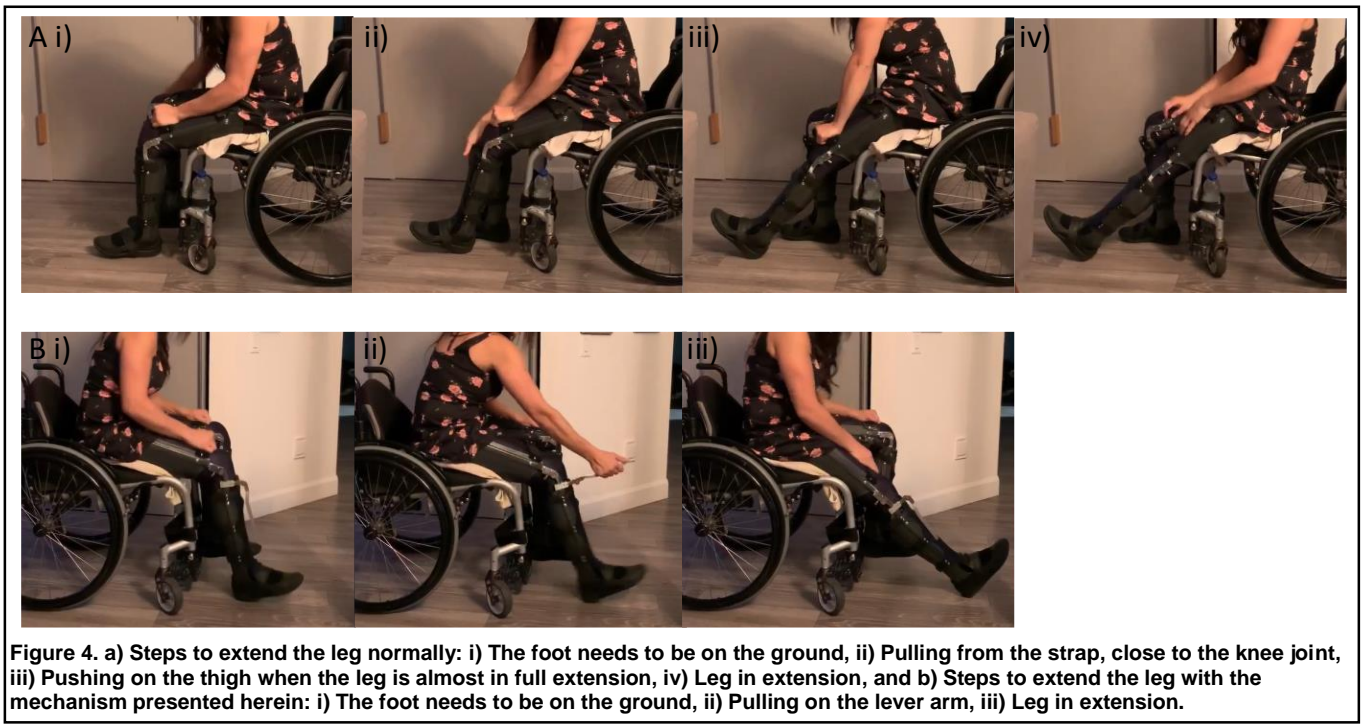

**Figure 4. a) Steps to extend the leg normally: i) The foot needs to be on the ground, ii) Pulling from the strap, close to the knee joint, iii) Pushing on the thigh when the leg is almost in full extension, iv) Leg in extension, and b) Steps to extend the leg with the mechanism presented herein: i) The foot needs to be on the ground, ii) Pulling on the lever arm, iii) Leg in extension.**

Figure 4 shows the difference to extend the leg without (a) and with (b) the mechanism. The mechanism in Figure 4b is faster and easier for the user.

## DISCUSSION

The mechanism allows the user to lock the joint of the orthosis while transforming a mainly parallel force to a mainly perpendicular force at the pivot point at the knee. A preliminary test indicates that the objective was met for this person but should be tester with other orthosis users.

The mechanism is designed for a knee-ankle-foot orthosis, but could easily be transferred to any other type of mechanical orthosis. Minor modifications can be made to the design to fit on other types of orthoses.

Future versions of the mechanism could include a means to adjust the anchor on the orthosis in the transverse plane. Currently, the user can only adjust the orthosis in the sagittal plane. This latter modification would allow the mechanism to fit on more complex types of orthoses.

The design is promising for persons who use mechanical orthoses to extend their knees in order to stand and walk. The mechanism requires less strength and the design makes it small and simple to use.

## CONCLUSION

This paper has presented a mechanism designed to help people using mechanical orthosis to extend their legs easily. The mechanism includes a bar that can help users extend their knees with less force. The complete design is foldable and therefore not cumbersome. The objective was to help users extend their leg with a minor modification to the design of the orthosis mechanism. In the short term, future work includes the adaptation of the mechanism to other types of mechanical orthoses.

## ACKNOWLEDGEMENT

This work is supported by Dr. Campeau-Lecours's internal funds at Université Laval.